\documentclass[manuscript]{acmart}
\AtBeginDocument{%
  }

\setcopyright{none}
\copyrightyear{2027}
\acmYear{2027}
\acmDOI{}
\acmConference[CHI '27]{CHI Conference on Human Factors in Computing Systems}{May 10--14, 2027}{Pittsburgh, PA, USA}
\renewcommand\footnotetextcopyrightpermission[1]{}

\begin{document}

\title[Configuration-aware human intervention boundaries]{CALM: Configuration-Aware Human Intervention Boundaries During Robot Approach}

\author{Xinting Gao}
\affiliation{%
  \institution{School of Architecture, Tsinghua University}
  \city{Beijing}
  \postcode{100084}
  \country{China}}

\author{Sipu Zhu}
\affiliation{%
  \institution{School of Architecture, Tsinghua University}
  \city{Beijing}
  \postcode{100084}
  \country{China}}

\author{Weimin Zhuang}
\correspondingauthor
\email{zhuangwm@tsinghua.edu.cn}
\affiliation{%
  \institution{School of Architecture, Tsinghua University}
  \city{Beijing}
  \postcode{100084}
  \country{China}}
\affiliation{%
  \institution{Architectural Design \& Research Institute of Tsinghua University}
  \city{Beijing}
  \postcode{100084}
  \country{China}}

\renewcommand{\shortauthors}{Gao et al.}

\begin{abstract}
How robot body configuration shapes human intervention during approach
remains underexplored. We conducted a within-participants study with
41 participants, measuring final stopping distance, subjective comfort,
and exploratory eye-tracking responses across four humanoid arm
configurations and two spatial scales. Full forward arm extension
increased stopping distance by approximately 31--36~cm relative to
arms-down. Spatial scale primarily affected comfort and pupil responses
without a detectable stopping-distance shift. We introduce the
Configuration-Aware Limit Model (CALM), which translates stopping-distance
distributions into configuration-dependent population-coverage boundaries.
Estimated boundaries at 80\% coverage ranged from 0.88 to 1.47~m.
In an illustrative one-dimensional planning analysis, reconfiguration
enabled a 1.10~m approach goal that was unreachable with arms remaining
fully extended under the same nominal pointwise 20\% intervention-probability
constraint. These findings support treating body configuration as a
planning variable while distinguishing physical safety, behavioral
intervention, and subjective cost.
\end{abstract}



\begin{teaserfigure}
  \centering
  \includegraphics[width=\textwidth]{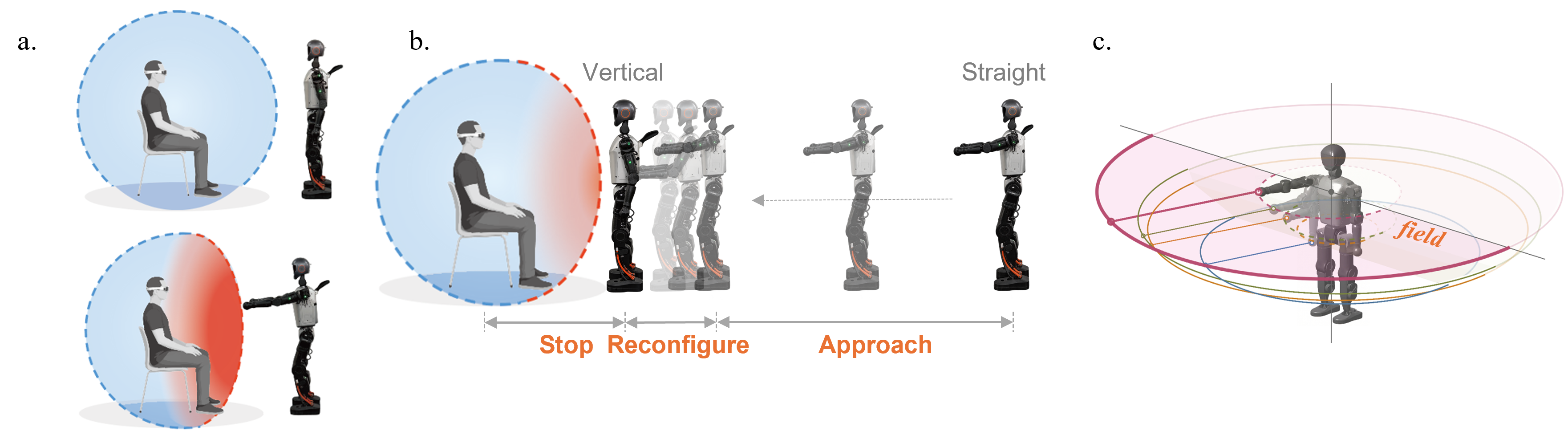}
  \caption{Study overview.
  (a) Arm configuration changes the robot's forward envelope and the human intervention boundary during frontal approach.
  (b) CALM-reconfig allows the robot to stop at a configuration-specific boundary,
  reconfigure from Straight to Vertical, and continue toward a closer goal under
  the same nominal pointwise intervention-probability constraint.
  (c) These configuration-dependent boundaries can be represented as a
  robot-centered spatial field for planning.}
  \Description{Study overview showing configuration-dependent human intervention
  boundaries, a CALM-based robot reconfiguration sequence, and a robot-centered
  representation of configuration-dependent spatial boundaries.}
  \label{fig:overview}
\end{teaserfigure}

\maketitle


\section{Introduction}

Robots are increasingly moving beyond isolated automation and operating in indoor environments that remain occupied by people, including offices, hospitals, public buildings, and other shared spaces \citep{lasotaSurveyMethodsSafe2017,paraschoConstructionRoboticsAutomation2023}. A defining feature of robots is their physical embodiment: unlike purely virtual agents, they occupy physical space and can act on and interact with the real world \citep{doi:10.1126/scirobotics.aed4569}. This embodiment makes spatial interaction particularly consequential when robots operate in close proximity to people. In tasks such as delivery, guidance, object handover, and collaboration, robots must not only avoid physical collisions but also approach people in ways that are acceptable to them. An approach may satisfy physical safety requirements yet still be perceived as intrusive, uncomfortable, or difficult to interpret, potentially causing a person to stop, avoid, or otherwise interrupt the robot \citep{akalinTaxonomyFactorsInfluencing2023,haneyPerceivedSafetyHumanRobot2026}. Beyond physical safety limits, it is therefore important to understand when an approaching robot reaches a distance at which a person chooses to intervene. We refer to this behaviorally expressed distance as the \emph{human intervention boundary}.

Research on human--robot proxemics has established that human--robot spacing is not fixed, but varies with robot appearance, motion characteristics, approach direction, interaction context, and individual differences \citep{samarakoonReviewHumanRobot2022,takayamaInfluencesProxemicBehaviors2009}. Robot appearance and individual perception can further shape spatial preferences and interaction behavior \citep{gasteigerFactorsPersonalizationLocalization2023a}. However, most prior work has focused on relatively stable robot characteristics or on the behavior of the robot as a whole, while comparatively little attention has been paid to changes in the body configuration of the same robot.

This distinction is particularly important for humanoid robots and mobile manipulators, whose occupied space changes with arm posture, torso orientation, grasp configuration, and carried objects. Even when the robot base or torso remains at the same location, different body configurations can substantially change the spatial relationship between the robot and a nearby person. Importantly, a person's stopping decision may not depend only on the physical distance between the robot and the human. Different whole-body configurations may also alter how the robot's spatial extent, reaching potential, and approaching presence are perceived, thereby shifting the boundary at which a person chooses to intervene. The relevant quantity may therefore be the robot's overall configuration rather than geometric distance or arm-extension length alone.

Whole-body planning already accounts for configuration-dependent geometry when evaluating collision and motion feasibility \citep{kennel-maushartPayloadAwareTrajectoryOptimisation2024,wuRealtimeWholebodyMotion2024}, whereas human-aware navigation more commonly represents human spatial preferences through personal-space fields or social costs \citep{eiraleLearningSocialCost2024a,martiniAdaptiveSocialForce2024}. A direct empirical connection between robot body configuration, human perception of an approaching robot, and behavioral intervention boundaries is still missing.

The surrounding spatial environment may further shape human responses to robot approach. Studies of robot passing and navigation in narrow or work-like environments have shown that available space influences movement coordination and how people respond to robot motion \citep{fujiokaNeedPassUnderstandable2024a}. Spatial context, however, may not affect all human responses in the same way. Stopping distance reflects an explicit and thresholded behavioral decision, whereas subjective comfort captures conscious evaluations of the interaction. Pupil-based measures can additionally provide information about visual attention, arousal, and processing demand \citep{bradleyPupilMeasureEmotional2008}. A person may therefore experience increased discomfort or monitoring demand without necessarily changing the final position at which they stop the robot. Behavioral intervention, subjective experience, and oculomotor response should consequently be examined as related but distinct aspects of human response.

Despite these advances, several questions remain unresolved. First, it is unclear whether different body configurations of the same humanoid robot systematically shift the boundary at which people intervene during a dynamic approach, beyond what would be expected from physical distance alone. Second, it remains unclear whether spatial scale modifies this configuration effect or primarily influences subjective and oculomotor responses. More broadly, even when such differences can be identified experimentally, there remains a gap between empirical human-response evidence and representations that can be directly incorporated into configuration-aware robot planning.

To address these questions, we conducted a controlled frontal-approach experiment in which a humanoid robot approached seated participants while systematically varying its arm configuration and the spatial scale of the environment. We examined stopping distance as the primary behavioral outcome and additionally measured subjective comfort and eye-tracking responses. We further introduce the \emph{Configuration-Aware Limit Model} (CALM), which transforms the observed stopping-distance distribution into configuration-dependent population-coverage boundaries. CALM represents human intervention risk rather than collision safety and is therefore intended to complement, rather than replace, physical safety constraints \citep{karagiannisAdaptiveSpeedSeparation2022,lasotaSurveyMethodsSafe2017}.

Accordingly, this study addresses three research questions:

\textbf{RQ1.} How does humanoid arm configuration affect the intervention boundary and subjective comfort during frontal robot approach?

\textbf{RQ2.} How does spatial scale affect behavioral, subjective, and oculomotor responses, and does it moderate the effect of robot configuration?

\textbf{RQ3.} Can stopping behavior be modeled as a configuration-dependent population-level intervention boundary, and does reconfiguration improve close-approach feasibility?

This work makes three contributions. First, it quantifies how different arm configurations of the same humanoid robot alter human intervention boundaries under otherwise controlled approach conditions, showing that stopping behavior cannot be understood solely from a fixed human--robot distance. Second, it distinguishes behavioral stopping, subjective comfort, and oculomotor responses, enabling robot configuration and spatial scale to be examined across different levels of human response. Third, it introduces CALM as a configuration-dependent population-coverage model and uses an illustrative planning analysis to show that reconfiguration can recover approach-goal feasibility without relaxing the nominal intervention-probability constraint.

\section{Related Work}

\subsection{Human--Robot Approach and Intervention Boundaries}

Human--robot approach distance has been studied from both
physical-safety and human-centered perspectives. Safe-HRI research
emphasizes collision avoidance, speed and separation monitoring, and
safety-aware control
\citep{lasotaSurveyMethodsSafe2017,
karagiannisAdaptiveSpeedSeparation2022}.
Existing standards define important but application-specific
physical-safety requirements. ISO~3691-4 addresses driverless
industrial trucks, whereas ISO~10218-2 and ISO/TS~15066 primarily
concern industrial robot applications and collaborative industrial
robot systems, respectively
\citep{iso3691_4_2023,iso10218_2_2025,iso15066_2016}.
Although their principles are informative for hazard reduction,
protective separation, and control reliability, these standards do
not directly define socially acceptable intervention boundaries for
humanoid robots operating in public or service environments. Within
their intended application contexts, they address physical hazards
and protective measures, but not the point at which a physically
permissible approach becomes unacceptable to a person or triggers
active intervention.

Human--robot proxemics therefore focuses on how people regulate spatial relationships with embodied robots. Dautenhahn et al.\ examined a robot approaching a seated person and showed that approach behavior affects accepted distance \citep{dautenhahnHowMayServe2006}. Takayama and Pantofaru found that proxemic behavior varies with characteristics of both the robot and the human \citep{takayamaInfluencesProxemicBehaviors2009}. Walters et al.\ further demonstrated effects of robot appearance and voice on approach distance \citep{waltersCloseEncountersSpatial2005,waltersHumanApproachDistances2008}, while Syrdal et al.\ highlighted substantial individual differences in spatial preferences \citep{syrdalPersonalizedRobotCompanion2007}. Robot appearance and individual perception have likewise been identified as important factors in personalized HRI \citep{gasteigerFactorsPersonalizationLocalization2023a}. A recent review of trust assessment further shows that proxemics is relevant across social-care and industrial settings, with human and environmental factors becoming increasingly important during interaction \citep{campagnaSystematicReviewTrust2025}.

Together, these studies show that human--robot spacing cannot be represented by a universal social radius. However, most prior work varies robot appearance, behavior, or interaction context while treating the robot body itself as relatively stable. For a humanoid robot, the same base position can correspond to substantially different body envelopes and perceived reaching potential as its arms reconfigure. People may therefore respond to the robot's overall body configuration rather than to physical distance alone. Whether such configuration changes systematically shift the point at which people actively stop an approaching humanoid robot remains underexplored. We refer to this behaviorally expressed threshold as the \emph{human intervention boundary}.

\subsection{Spatial Context and Multi-Component Human Responses}

Human--robot interaction is embedded in a physical environment whose geometry can shape both movement and perception. Fujioka et al.\ showed that understandable robot behavior is important for coordinating passing interactions in narrow environments \citep{fujiokaNeedPassUnderstandable2024a}. In work-like settings, Kim et al.\ found that people systematically adjust their trajectories and proxemic behavior around robots \citep{kimUnderstandingHumanRobotProxemic2024}, while related work showed that human-like robot motion can also alter responses during unfocused interaction \citep{kimUnderstandingEffectsHumanlike2025}. These findings indicate that spatial layout and robot behavior jointly influence how people negotiate shared space.

Existing studies have mainly focused on passing, trajectory adaptation, or mobile coordination. Less is known about whether spatial scale changes the intervention boundary of a stationary person during frontal robot approach. A constrained environment may increase perceived confinement, reduce perceived controllability, or heighten monitoring without necessarily shifting the final position at which a person decides to stop the robot. Spatial context may therefore affect subjective or perceptual responses even when the behavioral stopping boundary remains unchanged.

Different measures can capture these response levels. Bradley et al.\ linked pupil dilation to emotional arousal and autonomic activation \citep{bradleyPupilMeasureEmotional2008}, while Geng et al.\ showed that pupil diameter also varies with attentional uncertainty \citep{gengPupilDiameterReflects2015}. In HRI, Marchesi et al.\ used pupil dilation to characterize sensitivity to human-like cues in robot behavior \citep{marchesiLookingMindPupil2021}. Pupil-based measures should therefore be interpreted as indicators of attention, arousal, or processing demand rather than direct measures of safety. In the present work, stopping distance captures explicit behavioral intervention, subjective ratings capture conscious evaluation, and eye-tracking provides complementary evidence of oculomotor response. Examining these measures together allows us to test whether robot configuration and spatial scale affect different components of human response.

\subsection{Configuration-Dependent Robot Representation and Human-Aware Planning}

Robot planning research has increasingly recognized that articulated systems cannot always be represented by a fixed mobile-base geometry. Wu et al.\ proposed a whole-body planning framework that combines environment-adaptive search with spatial-temporal optimization for mobile manipulators \citep{wuRealtimeWholebodyMotion2024}. Kennel-Maushart and Coros further showed that payload and robot configuration directly affect trajectory feasibility and stability in mobile manipulation \citep{kennel-maushartPayloadAwareTrajectoryOptimisation2024}. These studies demonstrate that the occupied space and feasible motion of articulated robots are inherently configuration dependent. For human interaction, however, configuration changes may affect not only geometry but also how the robot's spatial extent and reaching potential are perceived.

Human-aware navigation addresses a complementary problem by incorporating human-related spatial preferences into motion planning. Eirale et al.\ learned social cost functions for human-aware path planning \citep{eiraleLearningSocialCost2024a}, while Martini et al.\ combined adaptive social-force representations with reinforcement learning for social navigation \citep{martiniAdaptiveSocialForce2024}. Samavi et al.\ used model predictive control and bilevel optimization for safe and interactive crowd navigation \citep{samaviSICNavSafeInteractive2025}, and Xu et al.\ incorporated uncertainty through distributionally robust chance-constrained trajectory optimization \citep{xuDistributionallyRobustChance2024a}. From the environment side, Gao et al.\ further treated spatial configuration as an optimization variable and jointly optimized environments and multi-agent trajectories through a differentiable bilevel formulation \citep{gaoDifferentiableEnvironmentTrajectoryCoOptimization2026}.

These directions nevertheless remain largely disconnected. Whole-body planning models configuration-dependent robot geometry, human-aware navigation models spatial preferences and interaction costs, and environment--trajectory co-optimization models how spatial configuration affects navigation performance. What is still missing is an empirical representation of how the robot's own body configuration changes human intervention behavior. Physical safety should remain an independent hard constraint determined by geometry, motion state, braking capability, sensing uncertainty, and control reliability \citep{pupaNovelDynamicMotion2025}. CALM instead targets the behavioral layer by estimating configuration-dependent population intervention boundaries from observed stopping-distance distributions, providing a link between human-response evidence and configuration-aware planning.

\section{Method}\label{sec:method}
\subsection{Participants}

Forty-one participants (16 men and 25 women; $M_{\mathrm{age}} = 24.73$ years, $SD = 5.25$, range = 18--44 years) were recruited from a university community through online recruitment platforms. Fourteen participants (34.1\%) reported prior direct interaction experience with physical robots, whereas 27 (65.9\%) reported no such experience. Detailed participant characteristics are provided in Supplementary Table~B1. G*Power \citep{faulGPower3Flexible2007} indicated that a minimum sample size of 34 participants was required to detect a medium-sized effect ($f = 0.25$) with $\alpha = 0.05$ and 80\% power for the within-subject repeated-measures design. All participants reported normal or corrected-to-normal visual acuity of 1.0 or better. All participants provided written informed consent and received a gift card equivalent to USD~15. The study was approved by the Institutional Review Board of [Anonymized University] (No.\ [Anonymized]).

\subsection{Robot Platform, Arm Configurations, and Experimental Spaces}

\begin{figure}[t]
    \centering
    \includegraphics[width=0.5\linewidth]{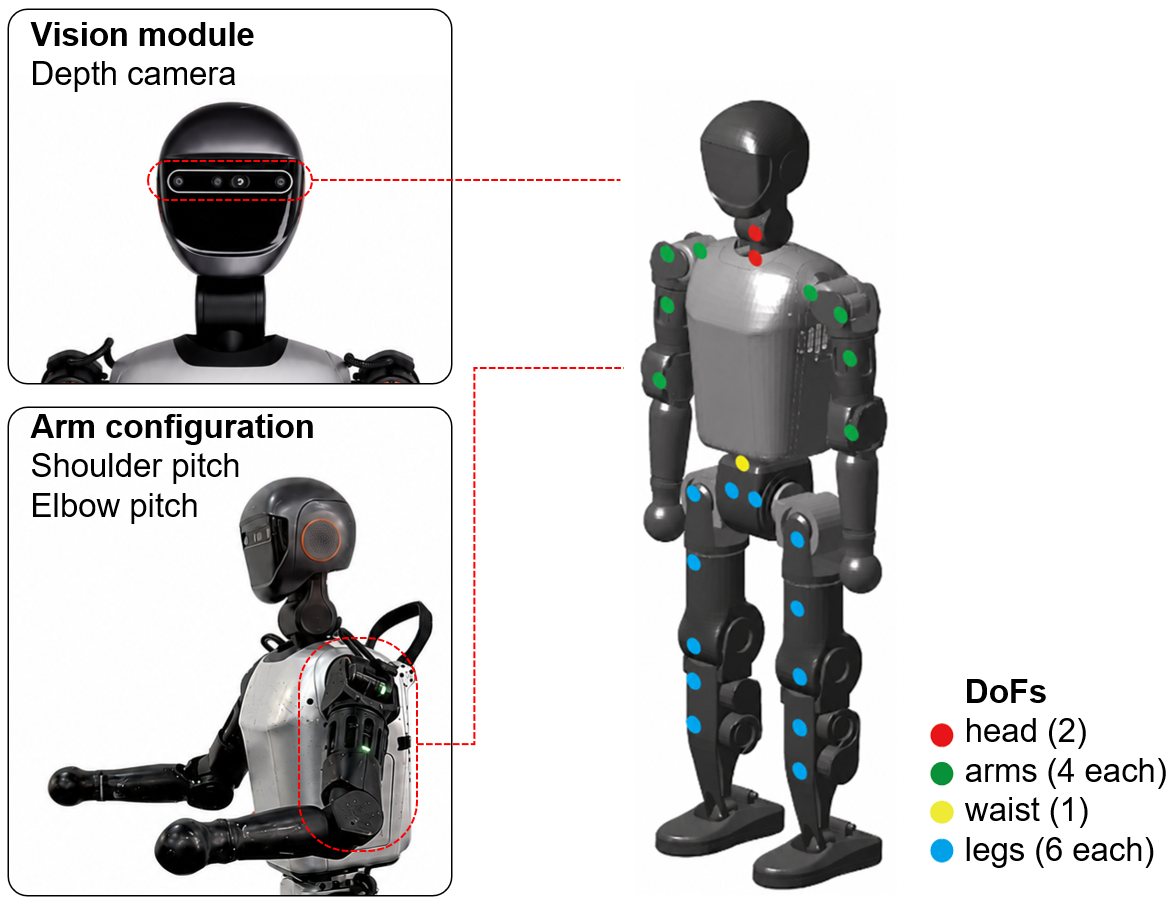}
    \caption{Experimental configuration of the Booster T1 humanoid robot, showing its degrees of freedom, onboard vision system, and manipulated arm configurations.}
    \label{fig:robotdetails}
\end{figure}

We used a Booster T1 humanoid robot (Booster Robotics, Beijing, China) \citep{boosterT1}. The robot is 1.18\,m tall, 47\,cm wide, weighs approximately 30\,kg, and has 23 degrees of freedom. Its onboard depth-camera system was used to estimate the human--robot distance during the approach (Figure~\ref{fig:robotdetails}). During the experiment, the robot operated in its built-in WALK mode at a commanded speed of 0.4\,m/s. Video-based verification showed that its actual walking speed remained stable across trials ($M = 0.40$\,m/s, $SD = 0.02$). The robot was controlled through the manufacturer's software development kit. Straight-line locomotion was generated using fixed velocity commands, while the arm configurations were produced using preprogrammed joint-angle targets. The same locomotion and stopping commands were used across all experimental conditions, and a hardware emergency stop remained available throughout the experiment.

\begin{figure}[t]
    \centering
    \includegraphics[width=\linewidth]{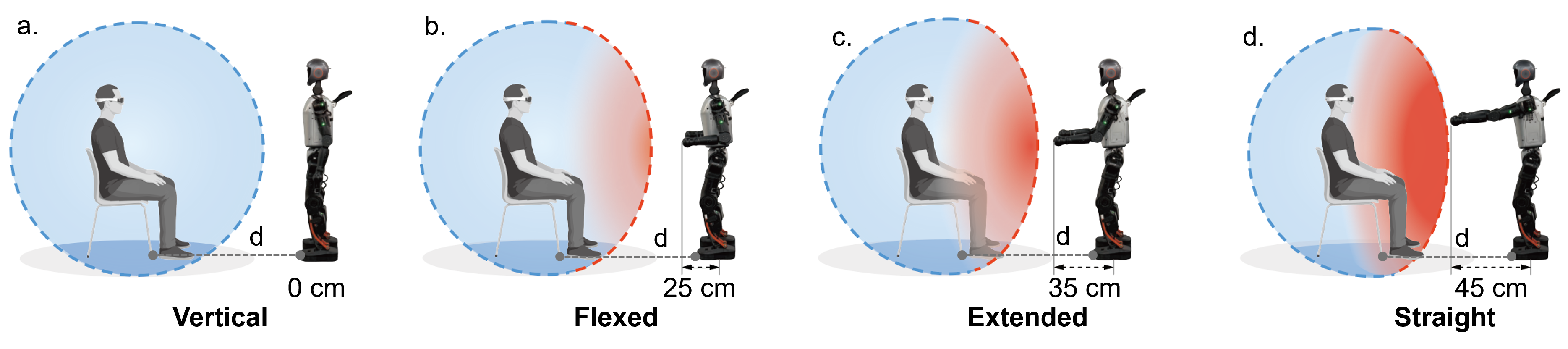}
    \caption{Arm extension as a manipulation of proxemic encroachment. Across the four postures (0, 25, 35, and 45\,cm of forward reach), body-to-body separation $d$ is held constant, while the extent to which the robot's forward body envelope encroaches on the participant's peripersonal space increases.}
    \label{fig:proxemics}
\end{figure}

The robot's forward body configuration was manipulated using four static arm postures (Figure~\ref{fig:proxemics}). Forward extension was defined as the horizontal distance from the anterior surface of the robot's torso to the hand endpoint, producing extensions of 0, 25, 35, and 45\,cm. Each posture was maintained throughout the approach while all other motion parameters were held constant. Thus, at a matched robot base position, increasing arm extension shifted the robot's forward body envelope toward the participant without altering its locomotion.

The experiment was conducted in two indoor spaces that differed in scale: Space A measured $2.5 \times 5.0$\,m and Space B measured $5.0 \times 5.0$\,m. In both spaces, the participant remained seated and the robot followed the same frontal approach path. A participant-centered coordinate system was established with the center of the participant's chair as the reference origin. At the beginning of each trial, the robot was positioned at an initial ground-plane distance of $d_0 = 3.5$\,m from this reference point. The robot's onboard depth-camera system was used to continuously estimate its position relative to the participant-centered reference frame during the approach. The relative positions of the participant, chair, robot starting point, and approach direction were matched across the two spaces. Both spaces were illuminated exclusively by artificial lighting at the same ambient illuminance, with no natural light present during data collection.

\subsection{Study Design and Procedure}
We conducted a $2 \times 4$ within-participants experiment to examine how spatial condition and robot arm configuration affected the stopping boundary adopted during a frontal humanoid--robot approach. The independent variables were space (Space A and Space B) and robot arm configuration (0, 25, 35, and 45\,cm of forward extension), yielding eight Space $\times$ Arm Configuration conditions that each participant completed in a participant-specific randomized order. The primary behavioral outcome was stopping distance; trial-level questionnaire responses and eye-tracking measures were additionally collected to characterize participants' subjective and oculomotor responses.

\begin{figure}[t]
    \centering
    \includegraphics[width=0.75\linewidth]{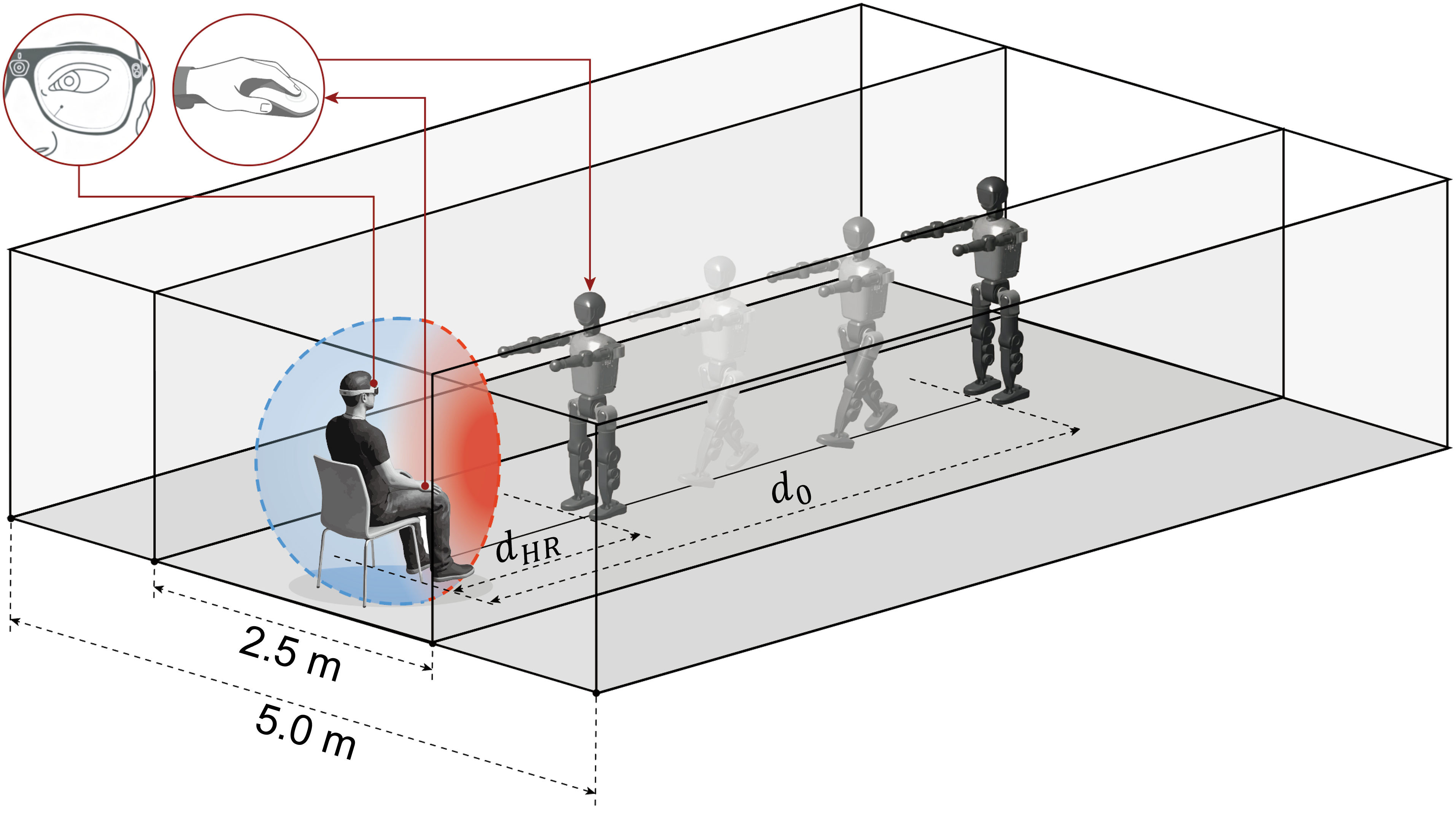}
    \caption{Experimental procedure for the robot approach task. The robot started at an initial distance $d_0$ from the participant and approached along a straight path. The participant used a mouse to stop the robot when the approach became uncomfortable. After stopping, the human--robot distance $d_{HR}$ was measured using the depth camera, while eye-tracking data were recorded throughout the approach. The procedure was repeated for four arm configurations under two spatial conditions.}
    \label{fig:procedure}
\end{figure}

Upon arrival, participants were seated at the predefined experimental position and first completed the calibration procedure for the eye-tracking system, followed by the pre-study questionnaire. The experimenter then explained the overall experimental procedure, familiarized participants with the task, and had them complete one practice trial. Resting-state eye-tracking data were subsequently recorded to establish an individual baseline for pupil-response correction. Before the formal trials began, participants were instructed to use the mouse button to stop the approaching robot when it reached the boundary of the distance they still considered comfortable. They then completed the eight experimental conditions, filling out the trial-level subjective questionnaire immediately after each one, and the post-study questionnaire after all conditions had been completed. Figure~\ref{fig:procedure} summarizes the experimental procedure.

\subsection{Measures}

At the beginning of each condition, the robot was initialized at the predefined starting distance $d_0 = 3.5\,\mathrm{m}$ and approached the participant along a straight path at $0.4\,\mathrm{m/s}$ while maintaining the assigned arm configuration. Participants were instructed to press a mouse button when they wanted the robot to stop. The button press triggered the robot's standard stopping command, and the final human--robot stopping distance, $d_{\mathrm{HR}}$, was recorded after the robot had fully stopped. Participants completed the trial-level questionnaire immediately after each condition. After completing all eight conditions, they completed a post-study questionnaire concerning their perceptions of the two spaces and the factors that influenced their stopping decisions. Figure~\ref{fig:procedure} summarizes the experimental procedure and the layouts of the two spaces.

\subsubsection{Stopping Distance}

Stopping distance, denoted by $D_{ij}$, was defined as the two-dimensional ground-plane distance between the robot and the participant after the robot had fully stopped, where $i$ denotes the participant and $j$ denotes the experimental condition. For each trial, a participant-centered coordinate system was established, with the center of the participant's chair calibrated as the origin. Because participants remained seated throughout the trial, this reference position remained fixed. The robot's onboard binocular depth camera was used to continuously estimate its two-dimensional position relative to this reference point. Stopping distance was calculated as $D_{ij}=\sqrt{x_{ij}^{2}+y_{ij}^{2}}$, where $x_{ij}$ and $y_{ij}$ denote the robot's planar coordinates relative to the participant-centered origin at the final stopping position. A larger $D_{ij}$ indicates that the participant stopped the robot at a greater distance. All analyses used the distance after the robot had fully stopped rather than the distance at the moment of the button press.

\subsubsection{Subjective Measures}

The questionnaire protocol consisted of pre-study, trial-level, and post-study components. The pre-study questionnaire collected participants' age, gender, and previous experience interacting with robots (Supplementary Appendix~A1).

Immediately after each experimental condition, participants completed a five-item task-specific questionnaire using seven-point Likert scales. Item development was conceptually informed by dimensions represented in the Godspeed Questionnaire \citep{bartneckMeasurementInstrumentsAnthropomorphism2009}, but the wording was tailored to the frontal-approach and proxemic context of the present experiment. The five items assessed perceived safety, postural intimidation,
personal-space intrusion, motion predictability, and perceived spatial
constraint. All items except motion predictability were negatively
worded and were reverse-scored, so that higher values consistently
indicated a more positive experience. The recoded scores were averaged
into a subjective comfort score, and internal consistency was assessed
using Cronbach's $\alpha$. The complete item wording is provided in
Supplementary Appendix~A2.

The post-study questionnaire asked participants to compare their perceptions of Space A and Space B and to report the factors that influenced their stopping decisions (Supplementary Appendix~A3).

\subsubsection{Eye Tracking}

Eye movements and pupil responses were recorded using a head-mounted Dikablis Glasses X eye tracker at a sampling rate of 120\,Hz. The eye-tracking system provided both pupil-diameter and pupil-area signals, and three measures were retained for analysis: Peak Pupil Diameter, Pupil Area Standard Deviation, and Mean Fixation Duration. Peak Pupil Diameter quantified the maximum task-evoked pupil response observed during the robot approach and was used as an index of peak arousal or cognitive processing demand \citep{joshiRelationshipsPupilDiameter2016}. Pupil Area Standard Deviation quantified the within-condition variability of the pupil-area signal \citep{reimerPupilFluctuationsTrack2014}. Mean Fixation Duration quantified the average duration of sustained visual processing during the approach \citep{zhangVisualAttentionCognitive2023} and was log-transformed before statistical analysis.

\subsection{Statistical Analysis}

The pupil-diameter and pupil-area signals were baseline-corrected against a
participant-specific baseline recorded before the formal trials. Gaps from
blinks or tracking loss were linearly interpolated up to 300\,ms; longer gaps
were left missing and excluded from the affected measures. For the exploratory time-resolved visualization, trials were aligned to
robot-movement onset and displayed over the first 7.5~s. A trial contributed
to a given time point only if that point fell within its actual recording
duration, without extrapolation beyond trial end. Because trial durations
varied, the set of contributing trials changed over time. Each measure was
accumulated from trial onset to the current time point, so the trajectories
represent cumulative rather than instantaneous responses and are interpreted
descriptively.

Stopping distance, subjective comfort scores, and the three eye-tracking measures were analyzed using separate linear mixed-effects models of the form

\begin{equation}
Y_{ij} \sim \text{Arm}_{j} \times \text{Space}_{j} + \left(1 \mid \text{Participant}_{i}\right),
\label{eq:lmm}
\end{equation}

with arm configuration (four levels) and space (two levels) entered as categorical fixed effects together with their interaction, and a by-participant random intercept included to account for repeated observations and individual differences.

Post-hoc pairwise comparisons were conducted using estimated marginal means, and $p$-values were adjusted for multiple comparisons using the Benjamini--Hochberg false discovery rate procedure \citep{benjaminiControllingFalseDiscovery1995}. Model diagnostics were used to examine residual normality and homoscedasticity. Fixed-effect estimates, 95\% confidence intervals, and FDR-adjusted $p$-values are reported.

\subsection{Configuration-Aware Limit Model (CALM)}

The Configuration-Aware Limit Model (CALM) complements the inferential analysis by translating the observed stopping distances into a population-coverage boundary that can be applied directly in robot motion planning. For a given arm configuration $A$ and spatial condition $S$, the boundary at coverage level $q$ is defined as the $q$th quantile of the population distribution of stopping distance and is estimated by the corresponding sample quantile:

\begin{equation}
\mathrm{CALM}_{q}(A,S) = F^{-1}_{D \mid A,S}(q),
\qquad
\widehat{\mathrm{CALM}}_{q}(A,S) = Q^{(7)}_{q}\!\left(D_{1}, \ldots, D_{n}\right),
\label{eq:calm}
\end{equation}

where $F_{D \mid A,S}$ is the population distribution of stopping distance under condition $(A,S)$, and $Q^{(7)}_{q}$ denotes the linearly interpolated sample $q$-quantile (Hyndman--Fan type~7 \citep{hyndmanSampleQuantilesStatistical1996}) computed over the $n$ participants in that condition.

CALM is defined for any coverage level $q$, and we report the 50th, 80th, and 
90th percentiles to characterize the boundary across a range of coverage 
requirements. We use $\mathrm{CALM}_{80}$ as the primary operating point 
throughout the paper; the remaining levels indicate how sensitive the boundary 
is to the chosen coverage level.

Uncertainty in the percentile boundaries was quantified using a participant-level cluster bootstrap. In each replicate, the boundary was recomputed with the same quantile estimator:

\begin{equation}
\widehat{\mathrm{CALM}}^{*(b)}_{q}(A,S)
= Q^{(7)}_{q}\!\left(D^{*(b)}_{1}, \ldots, D^{*(b)}_{n}\right),
\qquad b = 1, \ldots, B,
\label{eq:calm_boot}
\end{equation}

where $D^{*(b)}$ denotes the $b$th participant-level resample. Participants were drawn with replacement and all observations belonging to a sampled participant were retained together, so the same resampled participant set was used across all eight conditions. With $B = 5{,}000$, the 95\% confidence interval was taken as the 2.5th and 97.5th percentiles of the bootstrap estimates, again using the type~7 estimator.

The reported CALM values refer to the robot's final stopping position. A planner that issues the stop command in advance should offset the reported boundary by the braking displacement of its own platform.

\subsection{Illustrative CALM-Based Planning Analysis}
To examine how configuration-dependent boundaries affect approach decisions,
we conducted a one-dimensional planning analysis along the frontal approach
direction, with the robot starting in the Straight configuration at
$d_0 = 3.50$~m and CALM$_{80}$ as the nominal coverage level. Here,
$\widehat{P}_I(d,A)$ denotes the configuration-specific intervention-probability
value used by the planner at distance $d$ under arm configuration $A$.
Planning was subject to the pointwise constraint
$\max_t \widehat{P}_I(d_t,A_t) \leq 0.20$.

Three policies were compared. Fixed-clearance applied a single
configuration-invariant stopping distance of 0.88\,m, i.e. the most
permissive CALM$_{80}$ boundary applied irrespective of the current
configuration. CALM-fixed enforced the CALM$_{80}$ boundary of the
current configuration and remained in Straight throughout.
CALM-reconfig enforced the same constraint but could switch to another
experimental configuration after stopping; candidates that would
require the robot to retreat were excluded.

Planning used a 0.01~m grid, on which the executable boundary of a
configuration was defined as the first grid point satisfying
$\widehat{P}_I(d,A) \leq 0.20$
(0.89~m for Vertical, whose raw CALM$_{80}$ estimate was 0.88~m). The intervention-probability curves used for planning were evaluated
stepwise on the 0.01~m distance grid. Because these grid-based probability
values were evaluated separately from the Type-7 CALM quantile estimates,
the reported CALM$_{80}$ value need not coincide exactly with the first
grid point satisfying $\widehat{P}_I(d,A) \leq 0.20$.
Among admissible candidates the planner minimised
\begin{equation}
J(c)
=
\lambda_g |d_f - d_g|
+
\lambda_m |d_0 - d_f|
+
\lambda_c \mathbb{I}(c \neq c_0),
\label{eq:planning-objective}
\end{equation}
where $d_f$ is the final approach distance, $d_g$ the goal distance,
$c_0$ the initial configuration, and the $\lambda_m$ term penalises
unnecessary travel. We used $\lambda_g = 1.0$, $\lambda_m = 0.05$, and
$\lambda_c = 0.25$ (distances in metres). No separate minimum-gain
threshold was imposed: when both candidate stopping points lie beyond
the goal, these weights imply that a single configuration change is
preferred only if it reduces final goal error by more than
$\lambda_c/(\lambda_g-\lambda_m)=0.263$~m. Ties were broken by smaller
final error, then no configuration change, then smaller arm extension.

A goal counted as completed when the final grid point coincided with
$d_g$, i.e. zero final error on the 0.01\,m grid. A provisional lower
analysis bound of 0.80\,m was applied to all configurations; because
it was not derived from geometry-verified collision clearances, we
report task completion under the evaluated policies rather than
physical feasibility. Performance was evaluated by task completion,
final goal error, $\max_t\hat P_I$, and number of configuration
changes, for $d_g=1.10$\,m and for 17 goals from 0.80 to 1.60\,m in
0.05\,m steps.

\section{Results}\label{sec:results}

\subsection{Stopping Distance}
\label{sec:stopping_results}

\begin{figure}[h!]
  \centering
  \includegraphics[width=0.55\textwidth]{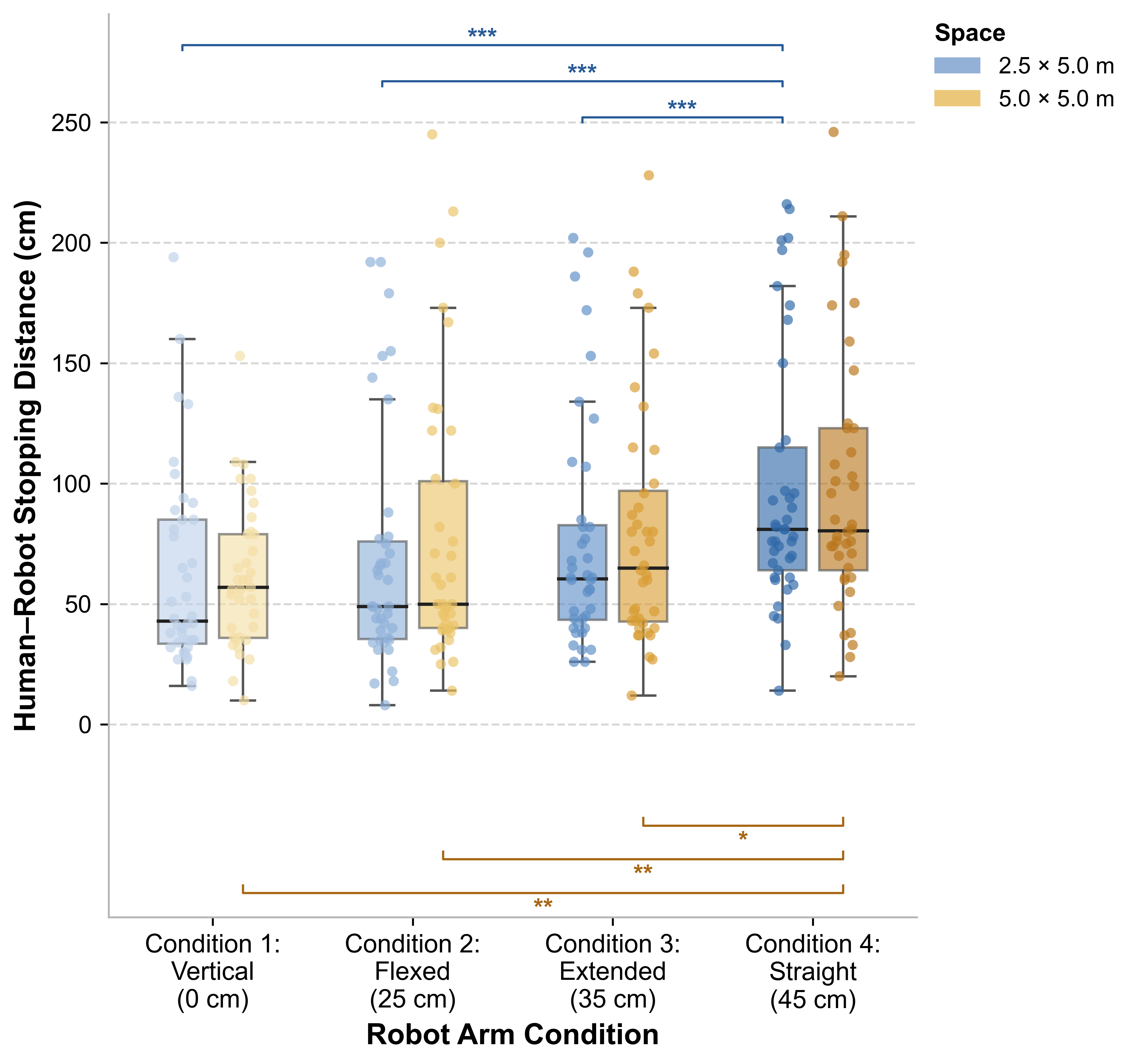}
  \caption{Stopping distance across robot arm configurations and spatial scales. Pairwise comparisons were FDR-corrected. Asterisks indicate statistical significance: * p < .05, ** p < .01, and *** p < .001.}
  \label{fig:distance}
\end{figure}

Arm configuration significantly shifted stopping distance, and this pattern was broadly consistent across the two spaces (Figure~\ref{fig:distance}). The mixed-effects model showed a significant main effect of arm configuration, $F(3,302)=36.65$, $p<.001$, whereas neither space, $F(1,302)=2.51$, $p=.114$, nor the Arm Configuration $\times$ Space interaction, $F(3,302)=0.55$, $p=.647$, was significant (Supplementary Table~B2).

In Space A, the estimated stopping distances for Vertical, Flexed, Extended, and Straight were 62.023 $\pm$ 7.550, 68.263 $\pm$ 7.572, 75.968 $\pm$ 7.550, and 98.098 $\pm$ 7.529 cm, respectively (EMM $\pm$ SE). Compared with Vertical, stopping distance increased by 13.95 cm for Extended ($p_{\mathrm{FDR}} = .003$) and by 36.07 cm for Straight ($p_{\mathrm{FDR}} < .001$). In Space B, the corresponding estimates were 66.461 $\pm$ 7.703, 76.265 $\pm$ 7.573, 79.262 $\pm$ 7.551, and 97.567 $\pm$ 7.551 cm. Compared with Vertical, stopping distance increased by 9.82 cm for Flexed ($p_{\mathrm{FDR}} = .043$), 12.82 cm for Extended ($p_{\mathrm{FDR}} = .008$), and 31.13 cm for Straight ($p_{\mathrm{FDR}} < .001$; Supplementary Tables~B3--B4).

Across both spaces, more forward-extending arm configurations were associated with larger stopping distances. The Straight configuration produced the largest shift. In contrast, spatial context affected subjective comfort and pupil responses but did not produce a statistically detectable shift in final stopping position.

\subsection{Subjective Evaluations}

Arm configuration was the main factor associated with subjective comfort, while space had a smaller but significant effect. The five trial-level questionnaire items showed high internal consistency (Cronbach's $\alpha = .898$) and were averaged into a subjective comfort score. Higher scores indicated greater comfort. The mixed-effects model showed significant main effects of arm configuration, $F(3,304)=48.91$, $p<.001$, and space, $F(1,304)=4.44$, $p=.036$, whereas the Arm Configuration $\times$ Space interaction was not significant, $F(3,304)=2.16$, $p=.093$ (Supplementary Table~B2).

\begin{figure}[h!]
  \centering
  \includegraphics[width=0.6\textwidth]{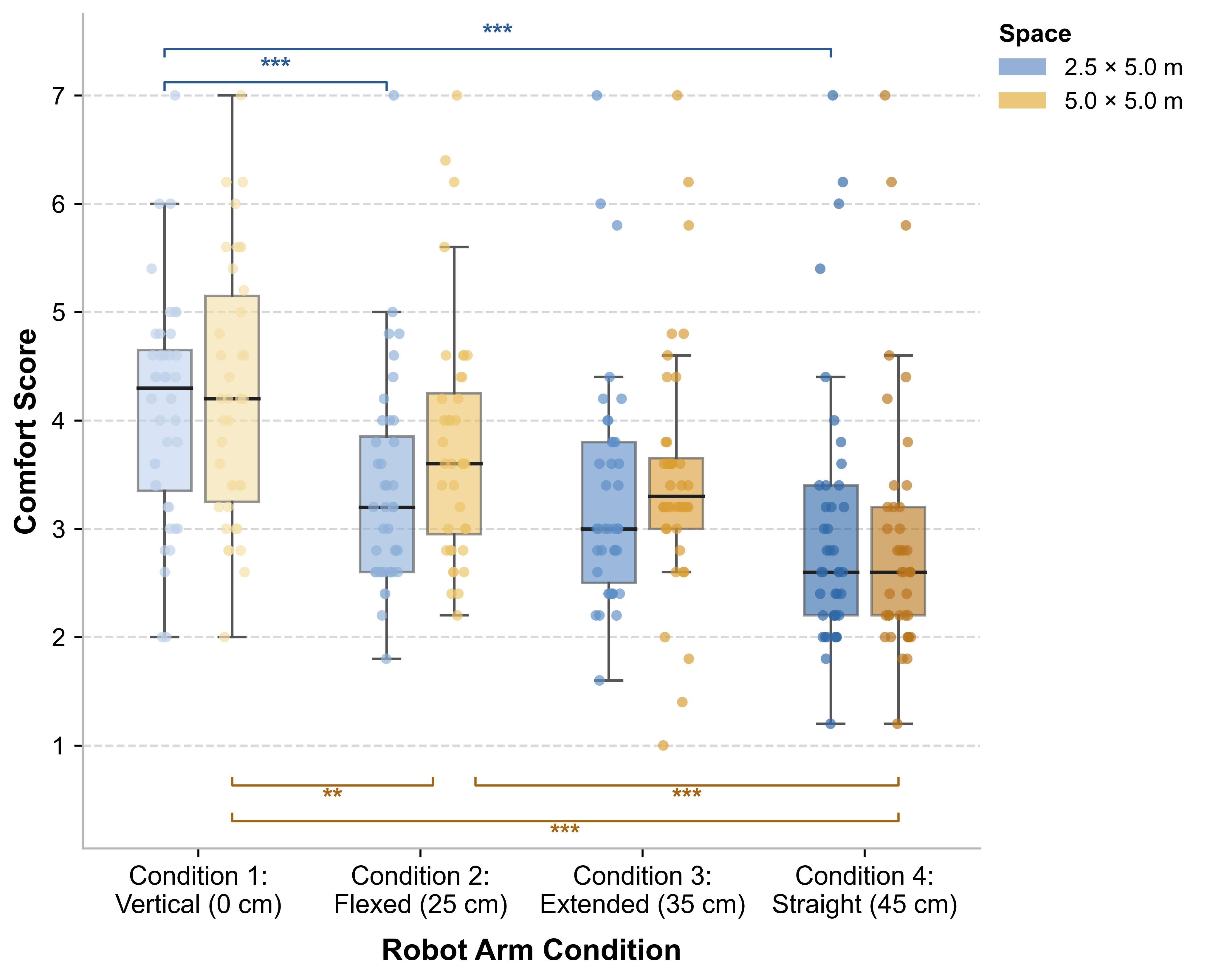}
  \caption{Distribution of participants’ comfort scores across two spatial conditions and four robot arm conditions. Asterisks indicate statistical significance: * p < .05, ** p < .01, and *** p < .001.}
  \label{fig:quationnare_mid}
\end{figure}

As shown in Figure~\ref{fig:quationnare_mid}, in Space A, the estimated comfort scores for the Vertical, Flexed, Extended, and Straight configurations were $4.116 \pm 0.170$, $3.331 \pm 0.171$, $3.356 \pm 0.170$, and $3.005 \pm 0.169$, respectively (EMM $\pm$ SE). Compared with Vertical, comfort was significantly lower for Flexed ($\Delta = 0.79$, $p_{\mathrm{FDR}} < .001$), Extended ($\Delta = 0.76$, $p_{\mathrm{FDR}} < .001$), and Straight ($\Delta = 1.11$, $p_{\mathrm{FDR}} < .001$). Space B showed the same overall pattern. The corresponding scores were $4.211 \pm 0.172$, $3.759 \pm 0.170$, $3.494 \pm 0.170$, and $2.939 \pm 0.170$. Compared with Vertical, comfort was lower for Flexed ($\Delta = 0.45$, $p_{\mathrm{FDR}} = .001$), Extended ($\Delta = 0.72$, $p_{\mathrm{FDR}} < .001$), and Straight ($\Delta = 1.27$, $p_{\mathrm{FDR}} < .001$; Supplementary Tables~B3--B4). Across both spaces, Vertical received the highest comfort ratings, whereas Straight received the lowest.

\begin{figure}[h!]
  \centering
  \includegraphics[width=0.8\textwidth]{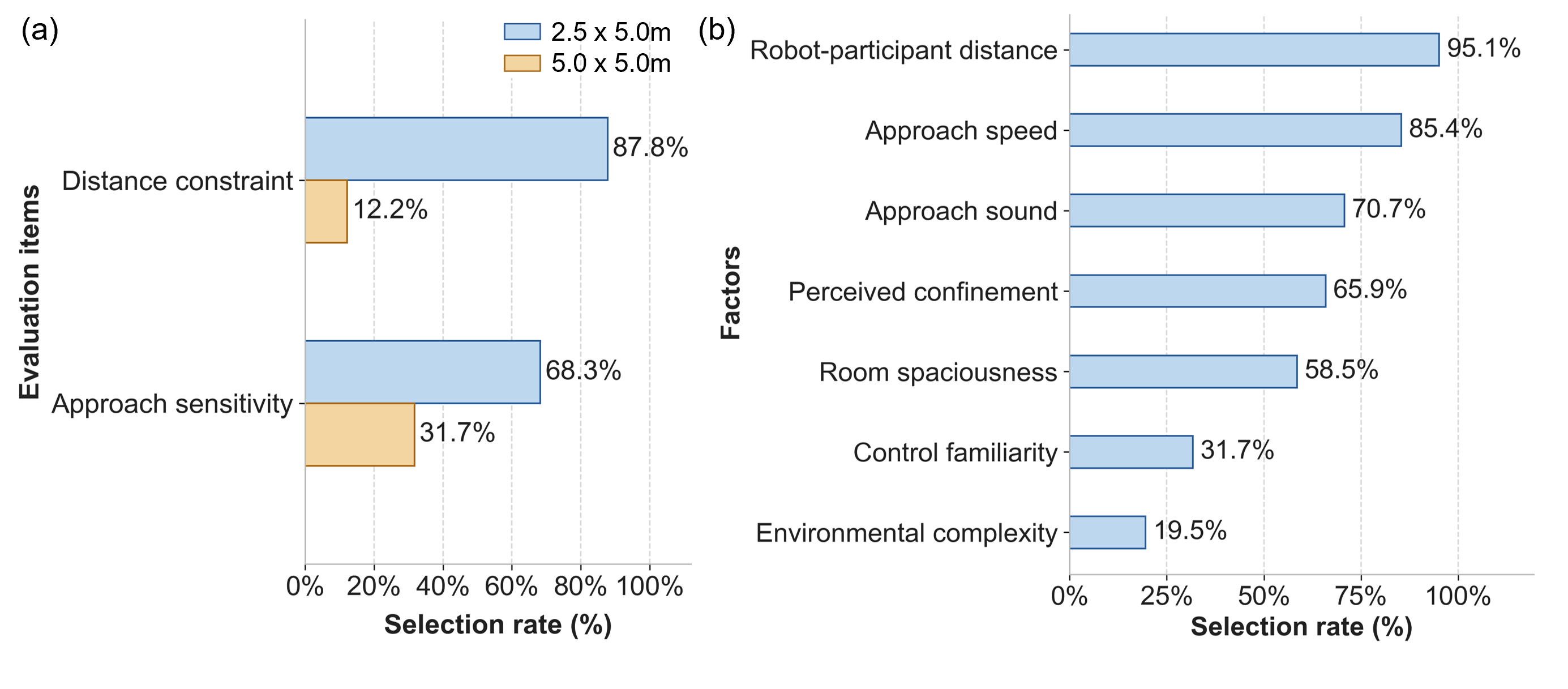}
  \caption{Participants' spatial perceptions and robot-stopping factors. (a) Perceived distance constraint and approach sensitivity across the two spatial conditions. (b) Factors influencing participants' robot-stopping decisions.}
  \label{fig:questionnare_after}
\end{figure}

Post-study responses showed that participants were also sensitive to the spatial context of the approach (Figure~\ref{fig:questionnare_after}). Thirty-six of 41 participants (87.8\%) reported feeling more constrained by the robot’s proximity in Space A, and 28 (68.3\%) reported being more sensitive to the robot's approach in that space. Human--robot distance was the most frequently selected factor affecting stopping decisions (39/41, 95.1\%). This was followed by approach speed (35/41, 85.4\%), approach sound (29/41, 70.7\%), perceived confinement (27/41, 65.9\%), and room spaciousness (24/41, 58.5\%). Familiarity with robot control (13/41, 31.7\%) and environmental complexity (8/41, 19.5\%) were selected less often.

\subsection{Exploratory Eye-Tracking Responses}
\label{sec:eye_results}

\begin{figure}[h!]
  \centering
  \includegraphics[width=1\textwidth]{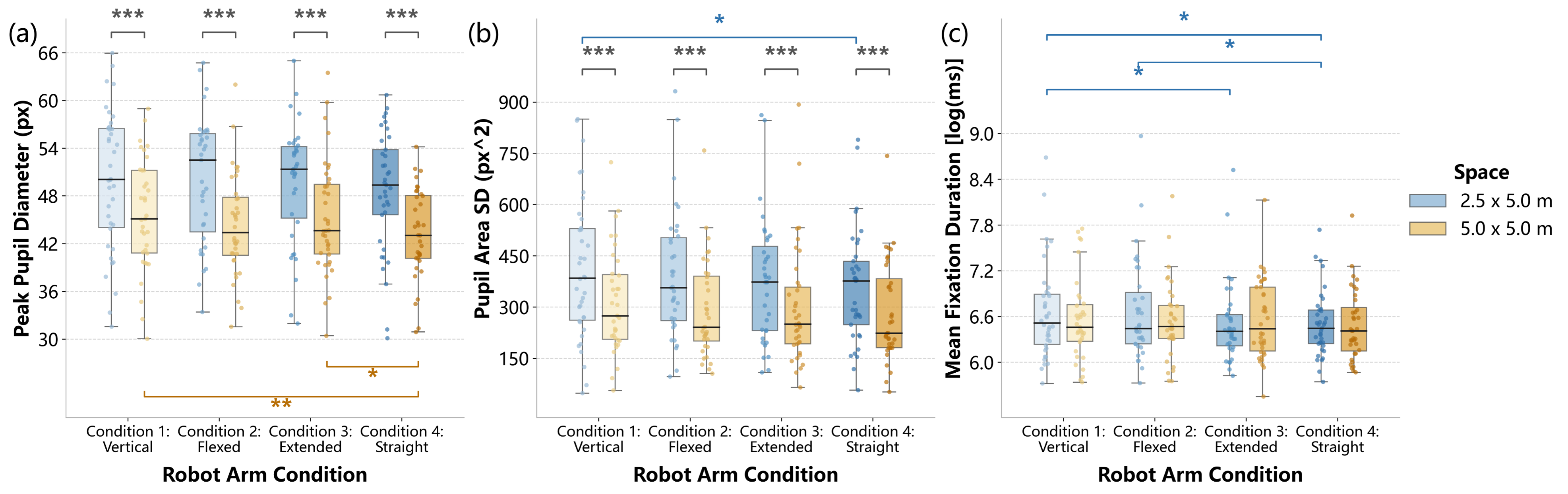}
  \caption{Statistical results for peak pupil diameter, pupil area SD, and mean fixation duration across two spatial conditions and four robot arm conditions. Asterisks indicate statistical significance: * p < .05, ** p < .01, and *** p < .001.}
  \label{fig:eye_box}
\end{figure}

Pupil-based measures mainly differentiated the two spatial conditions, while fixation duration showed weaker condition-dependent changes. For peak pupil diameter, arm configuration ($F=3.78$, $p=.011$) and space ($F=158.87$, $p<.001$) were significant, whereas their interaction was not ($F=0.45$; Supplementary Table~B2). Estimated peak pupil diameter ranged from $48.835 \pm 1.210$ to $50.352 \pm 1.219$\,px in Space A and from $43.382 \pm 1.215$ to $45.519 \pm 1.219$\,px in Space B (EMM $\pm$ SE; Supplementary Table~B3). Peak pupil diameter was significantly higher in Space A for all four arm configurations, with differences of 4.70--5.84\,px (all $p_{\mathrm{FDR}} < .001$). Within Space B, Vertical was also higher than Straight by 2.14\,px ($p_{\mathrm{FDR}} = .009$), and Extended was higher than Straight by 1.84\,px ($p_{\mathrm{FDR}} = .021$; Supplementary Table~B4; Figure~\ref{fig:eye_box}a).

Pupil area SD was primarily associated with space. Space showed a significant main effect ($F=75.56$, $p<.001$), whereas arm configuration ($F=2.24$) and the interaction ($F=0.39$) were not significant (Supplementary Table~B2). Estimated values ranged from $364.706 \pm 28.273$ to $412.603 \pm 28.161$\,px$^2$ in Space A and from $293.751 \pm 28.395$ to $316.894 \pm 28.508$\,px$^2$ in Space B (EMM $\pm$ SE; Supplementary Table~B3). Space A was significantly higher than Space B for all four configurations, with differences of 70.96--99.21\,px$^2$ (all $p_{\mathrm{FDR}} < .001$). An exploratory pairwise comparison also showed a higher value for Vertical than Straight in Space A ($\Delta = 47.90$\,px$^2$, $p_{\mathrm{FDR}} = .015$; Supplementary Table~B4; Figure~\ref{fig:eye_box}b).

Mean fixation duration showed no significant overall effect of arm configuration
($F = 2.26$), space ($F = 1.95$), or their interaction ($F = 1.24$;
Supplementary Table~B2). In Space A, exploratory pairwise comparisons showed
that Vertical was higher than Extended ($\Delta = 0.175$,
$p_{\mathrm{FDR}} = .032$) and Straight ($\Delta = 0.182$,
$p_{\mathrm{FDR}} = .024$). Flexed was also higher than Straight
($\Delta = 0.164$, $p_{\mathrm{FDR}} = .046$;
Supplementary Table~B4; Figure~7c). These comparisons are interpreted as local differences rather than evidence of a robust overall arm-configuration effect.

\begin{figure}[h!]
  \centering
  \includegraphics[width=1\textwidth]{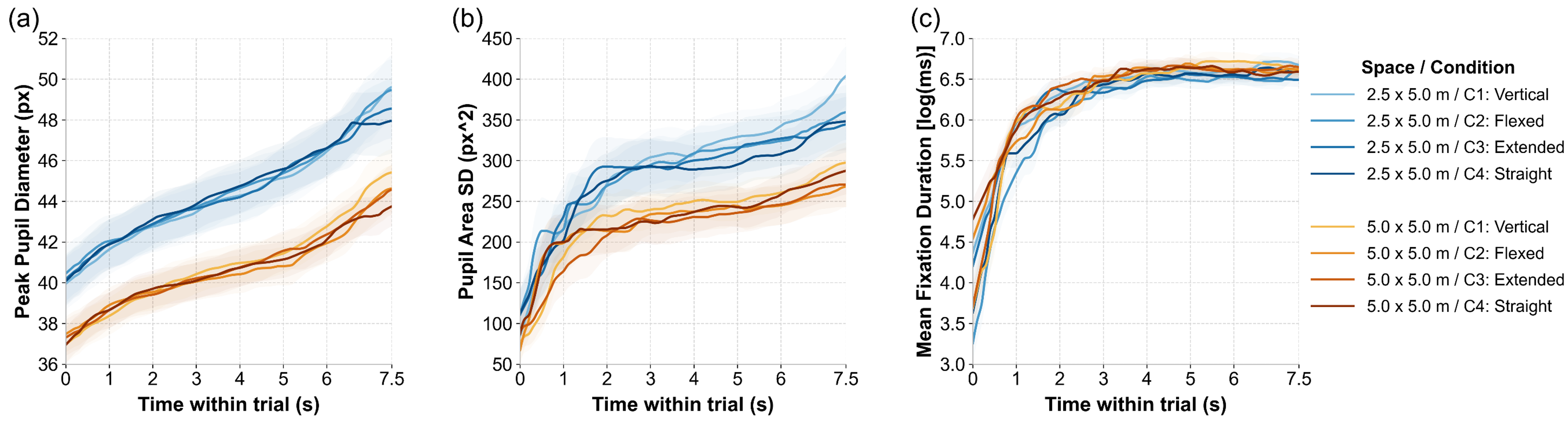}
  \caption{Exploratory cumulative trajectories of peak pupil diameter, pupil area SD, and mean fixation duration across two spatial conditions and four robot arm conditions. The curves are descriptive and are not used to estimate response onset.}
  \label{fig:eye_line}
\end{figure}

Figure~\ref{fig:eye_line} presents exploratory cumulative trajectories over the first 7.5\,s of the approach. Peak pupil diameter increased across the displayed window, with Space A remaining above Space B and substantial overlap among arm configurations within each space. Pupil area SD increased most rapidly during the first 1--2\,s, while mean fixation duration increased during the first few seconds and then approached a plateau. Because these measures were calculated cumulatively and the set of contributing trials changed as individual trials ended, the trajectories are interpreted descriptively rather than as evidence of response onset or an earlier physiological warning boundary.

\subsection{Configuration-Dependent CALM Boundaries}
\label{sec:calm_results}

CALM converted the observed stopping behavior into configuration-dependent population boundaries. Because neither the main effect of space nor the Arm Configuration $\times$ Space interaction was statistically detectable for stopping distance, we pooled observations across the two spaces to reduce sampling variability in the configuration-specific quantile estimates. This pooling was an estimation choice and should not be interpreted as evidence that the two spatial conditions were equivalent. We use $\mathrm{CALM}_{80}$, corresponding to 80\% population coverage, as the primary operating boundary. $\mathrm{CALM}_{50}$, $\mathrm{CALM}_{80}$, and $\mathrm{CALM}_{90}$ are reported in Supplementary Table~B5.

\begin{figure}[h!]
  \centering
  \includegraphics[width=0.7\textwidth]{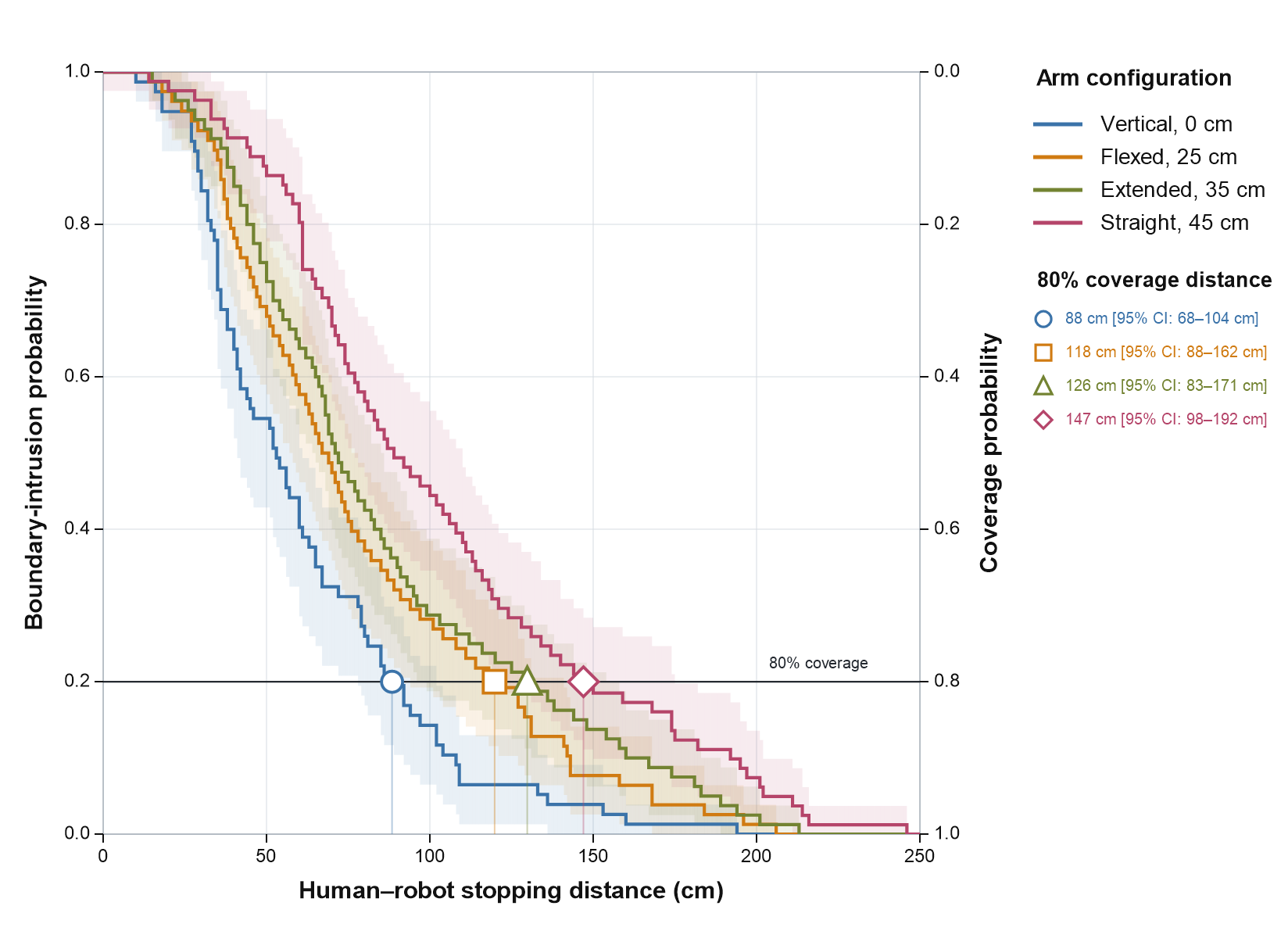}
  \caption{Configuration-Aware Limit Model (CALM) boundaries across four robot arm configurations. Step curves show the estimated boundary-intrusion probability as a function of human--robot stopping distance, with shaded areas representing 95\% confidence intervals. The right axis indicates the complementary coverage probability. The horizontal line marks 80\% coverage, equivalent to a boundary-intrusion probability of 0.20.}
  \label{fig:calm}
\end{figure}

As shown in Figure~\ref{fig:calm}, the boundaries of $\mathrm{CALM}_{80}$ were 88\,cm for Vertical, 118\,cm for Flexed, 126\,cm for Extended, and 147\,cm for Straight. Vertical produced the smallest boundary, while Straight produced the largest. This pattern was consistent with the stopping-distance results. In particular, the fully extended Straight configuration shifted the population stopping boundary outward. The boundaries did not scale linearly with geometric arm extension: the 25\,cm increase from Vertical to Flexed shifted the boundary by 30\,cm, whereas the subsequent 10\,cm increases to Extended and Straight shifted it by 8\,cm and 21\,cm, respectively.

\subsection{Approach Feasibility under CALM Constraints}
\label{sec:planning-implications}

Configuration-dependent CALM boundaries changed task-completion
outcomes under the evaluated policies. At $d_g=1.10$\,m,
Fixed-clearance reached the goal with zero final error but at
$\max_t\hat P_I=0.358$; its 0.88\,m clearance was not binding at this
goal distance. CALM-fixed held $\max_t\hat P_I\leq0.200$ but stopped
at the 1.47\,m Straight boundary, 0.37\,m short. CALM-reconfig
approached to 1.47\,m in Straight, retracted its arms to Vertical, and
continued to the goal, achieving zero final error under the same
constraint, with $\hat P_I=0.078$ on the final segment.

Across the 17 goals, CALM-reconfig completed 10 (58.8\%) versus 3
(17.6\%) for CALM-fixed: the 0.90--1.20\,m goals after switching to
Vertical, and the 1.50--1.60\,m goals while remaining in Straight, the
latter being the only goals CALM-fixed completed. The 0.80 and
0.85\,m goals were bounded by the executable Vertical boundary of
0.89\,m, leaving errors of 0.09 and 0.04\,m. For the 1.25--1.45\,m
goals, a more compact configuration would have satisfied the CALM
constraint, but the reduction in final error did not offset the
configuration-change cost.

\section{Discussion}\label{sec:discussion}

This study shows that arm configuration and spatial scale affected different components of human response during frontal humanoid approach. Forward-extending postures shifted the behavioral intervention boundary outward, whereas the narrower space primarily influenced subjective and oculomotor responses without a detectable shift in final stopping distance. CALM translated the stopping-distance distributions into configuration-dependent population-coverage boundaries, and the illustrative planning analysis showed that reconfiguration could recover a close-approach goal under an unchanged nominal predicted intervention-probability limit. Together, these findings support treating robot configuration as a planning variable while separating physical safety, behavioral intervention risk, and subjective cost.

\subsection{Configuration-Dependent Human Intervention Boundaries}

Robot configuration systematically shifted participants' intervention boundaries. Prior work has shown that human--robot spatial preferences vary with approach direction, motion characteristics, interaction context, prior experience, and individual differences \citep{samarakoonReviewHumanRobot2022,mummHumanrobotProxemicsPhysical2011a}. In the present study, compared with Vertical, Straight increased final stopping distance by 36.07 cm in Space A and 31.13 cm in Space B; Extended produced smaller increases of 13.95 cm and 12.82 cm, respectively. These effects indicate that a single configuration-independent distance is insufficient for representing human response to an articulated robot.

These center-to-center differences should not, however, be interpreted as purely psychological margins. Because stopping distance was measured from the robot center to the participant's chair centroid, the observed configuration effect combines the geometric change in the robot's forward body extent with any additional response to its overall posture or reaching potential. The present design cannot determine whether participants referenced the torso, the nearest forward point, or the robot's holistic configuration when deciding to intervene.

CALM should therefore be formulated as a configuration-aware probabilistic intervention model rather than as a fixed personal-space radius. It can represent the intervention probability associated with a given human--robot distance and body configuration, or the distance associated with a selected population-coverage level. This formulation places CALM at the behavioral-decision layer and avoids treating every increase in the robot's geometric extent as an equivalent increase in human spatial demand.

\begin{figure}[h!]
  \centering
  \includegraphics[width=0.75\textwidth]{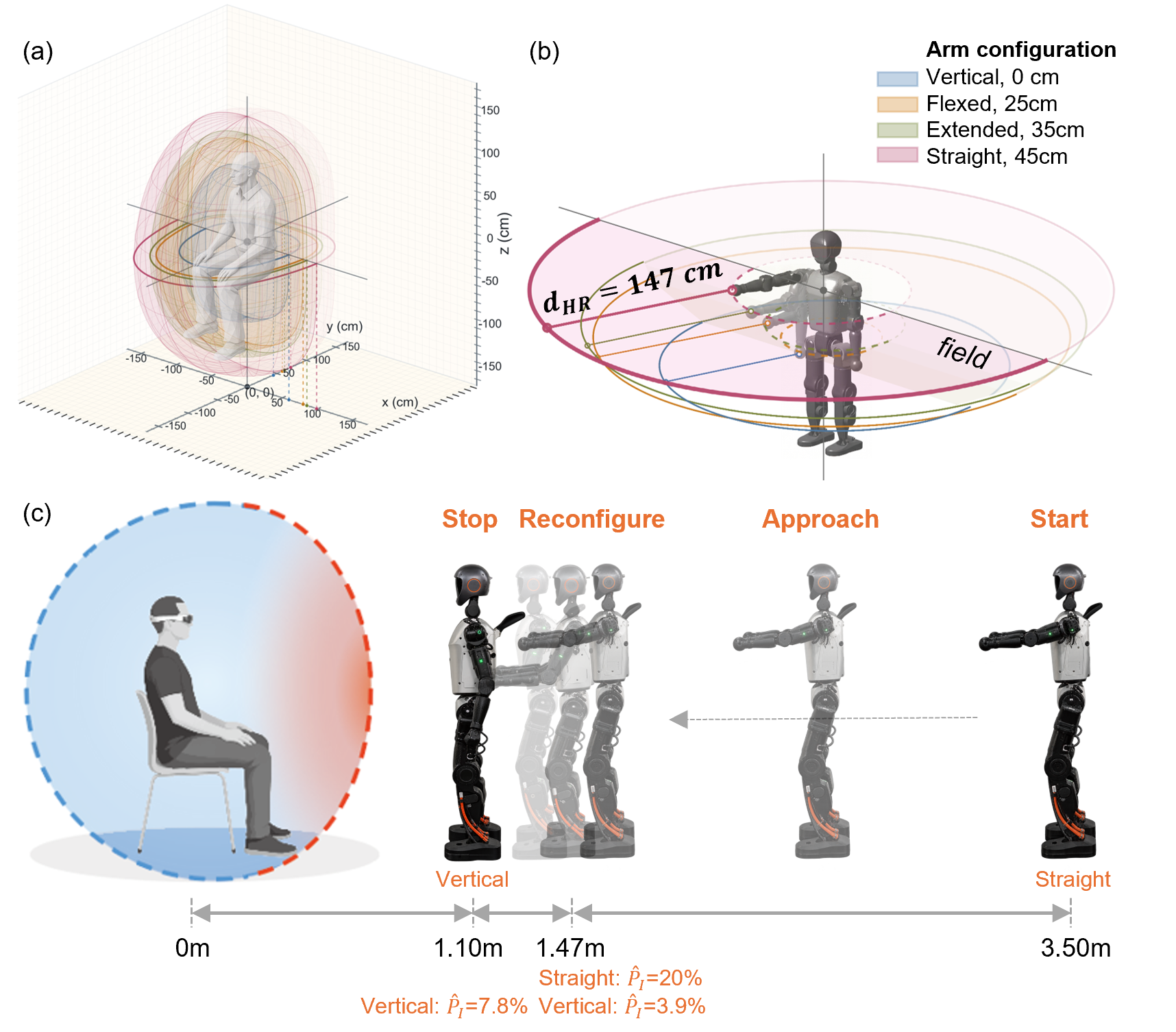}
  \caption{Configuration-dependent human--robot spatial boundaries. (a) Human-centered intervention boundaries across four arm configurations. (b) Robot-centered representation of the corresponding boundaries, indicating the configuration-dependent distance that the robot should maintain from a person. (c) Illustration of a CALM-based approach process. In the Straight configuration, the robot reaches the CALM$_{80}$ limit at 1.47\,m, where $\hat{P}_I = 20\%$. Reconfiguring to Vertical at the same distance reduces $\hat{P}_I$ to 3.9\%, allowing the robot to continue approaching. At the goal distance of 1.10\,m, $\hat{P}_I$ is 7.8\% ($\hat{P}_I$ denotes the predicted probability that a person will stop the robot when it is in a given configuration at distance $d$).}
  \label{fig:bubble}
\end{figure}

The directional boundary in Figure~\ref{fig:bubble}(a) is a conceptual extrapolation from the frontal-approach results. Only the frontal boundary was empirically calibrated because lateral and rear approaches were not tested. Following an asymmetric personal-space model, we set the normalized front, lateral, and rear extents to $1:0.67:0.50$ and used the measured frontal boundary to determine the overall scale \citep{Kirby-2010-10446}. The lateral and rear extents are therefore included only for visualization. Figure~\ref{fig:bubble}(b) conceptually extends the frontal findings into a robot-centered, configuration-dependent spatial field. Existing navigation systems combine robot footprint, obstacle distance, motion speed, and costmap inflation to represent collision risk and clearance \citep{luLayeredCostmapsContextsensitive2014b}, while social navigation encodes personal space and social constraints as navigation costs \citep{martiniAdaptiveSocialForce2024}. For articulated robots, the human-response field should additionally update with body configuration.

\subsection{Differential Effects of Spatial Scale Across Human Responses}

Spatial scale affected subjective and oculomotor responses more consistently than the final behavioral boundary. Space A did not produce a detectable change in final stopping distance, yet 87.8\% of participants judged the robot to be more spatially constrained there and 68.3\% reported greater sensitivity to its approach. Peak pupil diameter and pupil-area variability were also higher in Space A. Participants therefore registered the spatial manipulation even when it did not reliably shift the point at which they stopped the robot.

This divergence suggests that stopping distance reflects a thresholded intervention decision rather than a continuous measure of discomfort or monitoring demand. In the present task, the path, speed, and approach direction were fixed, and participants could respond only by pressing a stop button. Spatial pressure could not be expressed through lateral avoidance, path negotiation, yielding, or robot replanning, although these behaviors are central to interaction in constrained shared spaces \citep{fujiokaNeedPassUnderstandable2024a}. The absence of a detectable stopping-distance effect should therefore not be interpreted as evidence that the two spaces were behaviorally equivalent.

The measures may also capture different components or stages of human response. Eye measures primarily reflect visual attention and monitoring, questionnaires capture conscious evaluation, and stopping distance represents an explicit behavioral decision \citep{gaoEvaluatingImpactSpatial2025}. A robot may increase monitoring demand or subjective pressure before reaching the threshold for active intervention. This multi-component pattern also explains why the pooled CALM estimates should be understood as a choice made to improve quantile-estimation stability, rather than as evidence that spatial context is irrelevant.

\subsection{Implications for Configuration-Aware Planning}

Human-aware approach planning should distinguish three layers: physical safety, intervention risk, and subjective cost. Physical safety depends on body geometry, current configuration, speed, braking distance, control response, and sensing uncertainty, and should remain a hard constraint. Intervention risk concerns the probability that a person will stop, avoid, or otherwise interrupt the robot's motion; CALM primarily represents this layer. Subjective cost captures discomfort, pressure, or monitoring demand that may arise before intervention and is more appropriately represented as a soft cost.

\begin{figure}[h!]
  \centering
  \includegraphics[width=0.75\textwidth]{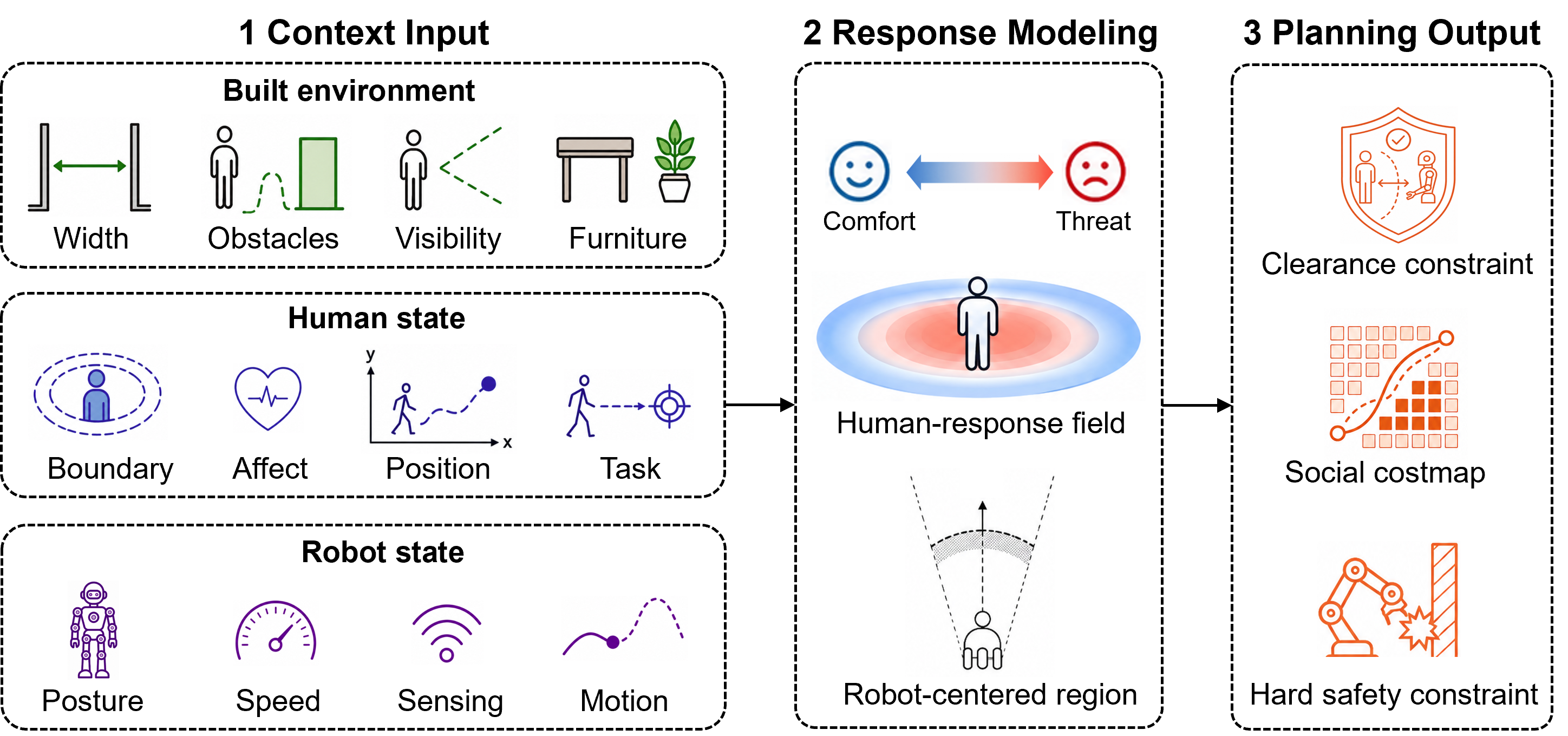}
  \caption{Mapping contextual inputs to human-response models and robot planning representations.}
  \label{fig:workflow}
\end{figure}

Figure~\ref{fig:workflow} maps these response layers to planning representations. Configuration-dependent intervention boundaries can be converted into clearance margins or chance constraints \citep{xuDistributionallyRobustChance2024a}. Subjective responses can be represented through a social costmap that shapes trajectory selection and motion profiles \citep{eiraleLearningSocialCost2024a}. Physical safety should remain an independent hard constraint based on the robot's motion state, stopping capability, and perception uncertainty \citep{pupaNovelDynamicMotion2025,karagiannisAdaptiveSpeedSeparation2022}. Keeping these layers separate prevents collision safety, behavioral intervention, and subjective discomfort from being collapsed into a single social radius.

Reconfiguration also recovered approach-goal feasibility without relaxing the nominal predicted intervention-probability constraint. For the 1.10~m goal, CALM-fixed reached the $\mathrm{CALM}_{80}$ boundary of the initial Straight configuration at 1.47~m and remained 0.37~m short of the goal. CALM-reconfig then retracted the arms to the Vertical configuration and continued to 1.10~m while maintaining the same constraint, $\max_t \hat{P}_I \leq 0.20$. Feasibility was therefore restored by changing the robot's body configuration---and consequently the applicable configuration-specific CALM boundary---rather than by accepting a higher predicted intervention probability. This result positions body configuration as a planning variable that can recover close-approach goals when a fixed configuration is infeasible. The illustrative analysis enforces a pointwise constraint on the estimated
stopping-boundary exceedance probability; it does not establish a 20\%
upper bound on intervention over an entire approach.

This planning result should be interpreted as an illustrative consequence of the empirically estimated boundaries rather than as validation of a complete reconfiguration planner, because configuration-transition dynamics and human responses to the reconfiguration itself were not evaluated. Nevertheless, it suggests that reconfiguration can be considered before modifying the base path. A robot might retract its arms, rotate its body, or change its grasp; for carrying tasks, payload and grasp configuration should also be included in whole-body planning and collision models \citep{wuRealtimeWholebodyMotion2024,kennel-maushartPayloadAwareTrajectoryOptimisation2024}. The robot could additionally slow down, shift laterally, wait, or adjust its orientation, while gaze, motion, light, or auditory cues may improve motion legibility \citep{angelopoulosYouAreMy2022,pascherHowCommunicateRobot2023}. The present study tested only one speed and static arm configurations, and therefore cannot establish how these adaptations interact.

CALM should complement physical safety standards rather than replace them. ISO~3691-4 \citep{iso3691_4_2023}, ISO~10218 \citep{iso10218_2_2025}, and ISO/TS~15066 \citep{iso15066_2016} provide application-specific requirements for physical hazards, protective separation, braking, and control reliability in industrial robot contexts. CALM instead estimates whether a physically permissible approach is likely to reach a person's behavioral intervention threshold. A deployed system should first satisfy the applicable physical safety requirements and then optimize intervention risk and subjective cost.

At this stage, $\mathrm{CALM}_{80}$ is better treated as a performance-based intervention-clearance metric than as a fixed building or safety standard. If $B_{\mathrm{human}}$ denotes the human intervention margin after subtracting the robot's configuration-dependent body extent, the current body extent can be added to this margin to obtain a total clearance. This formulation would avoid counting arm extension twice. However, the present data cannot independently separate body-envelope expansion from the additional human margin, and $\mathrm{CALM}_{80}$ therefore requires further calibration and validation.

\subsection{Limitations}

First, the restricted experimental conditions limit generalizability. We tested one humanoid robot, one speed of 0.4~m/s, four static arm configurations, and two spatial scales ($2.5 \times 5.0$~m and $5.0 \times 5.0$~m). Participants remained seated, and the robot followed a straight frontal path. The findings may not generalize to higher speeds, dynamic arm movements, moving people, other approach directions, carrying tasks, or crowded environments.

Second, the controlled button-press task and the measurement chain limit ecological validity and estimation precision. In natural settings, people may step backward, turn sideways, detour, slow down, speak to the robot, or show no visible response. Because distance was recorded only after the robot had fully stopped, the measured value does not isolate the participant's decision boundary at the moment of intervention. It combines the participant's decision and button-press timing with input and system-communication latency, platform braking displacement, controller execution error, and the final localization estimate. We also did not independently quantify depth-camera accuracy or trial-to-trial repositioning error. The reported boundary should therefore be interpreted as the final stopping outcome produced by this specific human--machine--control chain, rather than as a pure cognitive decision threshold.

Third, the sample composition limits population-level generalization. Participants were recruited mainly from a university community and largely represented an East Asian cultural context. Fourteen participants (34.1\%) had previously interacted with a physical robot, whereas 27 (65.9\%) had no such experience (Supplementary Table~B1). The sample was not sufficiently large or diverse to estimate interactions among culture, robot experience, personality, and personal-space preferences \citep{joosseCulturalDifferencesHow2014}. The reported distances should therefore not be interpreted as fixed thresholds for all populations.

\subsection{Future Work}

First, longitudinal studies should examine how intervention boundaries change with repeated exposure. Long-term HRI research shows that behavior, engagement, and perceptions of robots may change across repeated interactions \citep{matheusLongTermInteractionsSocial2025,labanBuildingLongTermHuman2024,abbasiLongitudinalStudyChild2025}. Future studies could track stopping distance, subjective evaluation, gaze, and physiological responses across days or weeks. This would allow CALM to account for habituation, familiarity, and cumulative experience, and could support online adaptation of population-level boundaries to individual users.

Second, CALM should be integrated into social navigation, whole-body planning, and closed-loop control. Its outputs could be implemented in ROS~2/Nav2 as a configuration-aware costmap layer, a local-planning cost, or a chance constraint \citep{xuDistributionallyRobustChance2024a}. At the control level, CALM could be combined with model predictive control and whole-body control to coordinate base motion, joint configuration, and speed \citep{duHierarchicalTaskModel2024,khazoomTailoringSolutionAccuracy2024}. Once calibrated, CALM could also provide an intervention-risk signal for reinforcement learning, serving as a reward term, constrained objective, or safety filter. A learned policy could select when to retract the arms, rotate, slow down, wait, or change path based on human--robot distance, current configuration, spatial context, and uncertainty \citep{martiniAdaptiveSocialForce2024}. Constrained or safe reinforcement learning would be preferable to unconstrained exploration because physical safety must remain an independent hard constraint. Training could begin with simulation or offline interaction data and then be evaluated through human-in-the-loop closed-loop studies.

Third, CALM should be validated across robot platforms and real-world contexts. Hospitals, stations, shopping centers, offices, and campuses could support studies of standing, walking, queueing, and multi-person interaction. These studies should examine the combined effects of robot type, task, speed, payload, dynamic configuration, and spatial layout. Cross-platform datasets would support model reproduction, out-of-sample calibration, and comparison with configuration-independent baselines. Only after validation across populations, platforms, and contexts should $\mathrm{CALM}_{80}$ be used as an operational parameter connecting robot design, spatial design, and deployment management.

\section{Conclusion}\label{sec:conclusion}

This study demonstrates that human intervention boundaries during frontal humanoid approach depend on robot arm configuration. Forward-extending postures shifted final stopping distances outward, whereas spatial scale primarily affected subjective and oculomotor responses without a detectable shift in the final behavioral boundary. CALM converts these stopping-distance distributions into configuration-specific population-coverage boundaries and represents predicted intervention risk rather than physical safety. In the illustrative planning analysis, reconfiguration recovered a 1.10~m approach goal under the same nominal constraint, $\max_t \hat{P}_I \leq 0.20$, whereas the fixed Straight configuration stopped at 1.47~m. These findings support treating body configuration as a planning variable while modeling physical safety, intervention risk, and subjective cost as separate layers. Future work should validate CALM across populations, platforms, speeds, dynamic configurations, and naturalistic interactions, and evaluate its out-of-sample calibration and closed-loop planning performance.

\begin{acks}
The authors gratefully acknowledge Booster Robotics for their support of this work.
\end{acks}

\bibliographystyle{ACM-Reference-Format}
\bibliography{main}

\end{document}